\pdfoutput=1
\documentclass[10pt]{llncs}
\usepackage{llncsdoc}
\usepackage[english]{babel}
\usepackage[utf8]{inputenc}
\usepackage[super]{nth}
\usepackage{booktabs}
\usepackage{amsmath}
\usepackage{amsmath,amssymb, graphicx}
\usepackage{lipsum}
\usepackage{romannum}
\usepackage{paralist} % also compact lists
\usepackage{multirow}
\usepackage{rotating}
\usepackage[bottom]{footmisc}
\usepackage{hyperref}
\usepackage[mathscr]{eucal} %% For \mathscr
\usepackage{amsbsy} %% For \boldsymbol
\usepackage{subfig}
\usepackage{booktabs}
\usepackage{xcolor}
\usepackage{tikz}
\usetikzlibrary{snakes}
\usepackage{multirow}
\usepackage{flushend}
\usepackage{algpseudocode}
\usepackage{algorithm2e}
\usepackage{tabularx}
\usepackage{epstopdf}
\usepackage{balance}
\usepackage{tablefootnote}
\usepackage{empheq} % added by christos - makes nice box for equations
\usepackage{moresize}
\usepackage{wrapfig}
\usepackage{lipsum}
\usepackage[inline]{enumitem}
\usepackage{soul,color}
\usepackage{chngpage}
\usepackage{calligra}
\usepackage[font=small,labelfont=bf,skip=0pt]{caption}
\usepackage{subfig}
\DeclareCaptionType{copyrightbox}
\usepackage{url}
\usepackage[figurename=Figure.]{caption}
\usepackage{xspace}

\newcommand{\reminder}[1]{{\textsf{\textcolor{red}  {[#1]}}}}

\newcommand{\hide}[1]{}
\newcommand{\ben}{\begin{enumerate*}}
\newcommand{\een}{\end{enumerate*}}
\newcommand{\bit}{\begin{itemize*}}
\newcommand{\eit}{\end{itemize*}}

\newcommand{\N}{\mathcal{N}}
\newcommand{\V}{\mathcal{V}}
\newcommand{\E}{\mathcal{E}}

\newcommand{\method}{\textsc{RECS}\xspace}

\begin{document}

\author{Saba A. Al-Sayouri$^1$, Danai Koutra$^3$, Evangelos E. Papalexakis$^2$, \\Sarah S. Lam$^1$}

\institute{State University of New York at Binghamton, Binghamton, USA\\
\email{\{ssyouri1, sarahlam\}@binghamton.edu}
\and
University of California Riverside, Riverside, USA\\
\email{epapalex@cs.ucr.edu }
\and
University of Michigan, Ann Arbor, USA\\
\email{dkoutra@umich.edu }}

\title{RECS: Robust Graph Embedding Using Connection Subgraphs}
\maketitle

%\keywords{information networks, network flow, learning graph representations, node embedding}
\begin{abstract}
% Graph embeddings have increasingly grown in popularity as node representations during the last years. Such representations encode the relations among distinct nodes in a continuous feature space. Further, they can generalize over a variety of tasks, such as node classification, link prediction, and recommendation.
The success of graph embeddings or node representation learning in a variety of downstream tasks, such as node classification, link prediction, and recommendation systems, has led to their popularity in recent years. 
Representation learning algorithms aim to preserve local and global network structure by identifying node neighborhood notions. However, many existing algorithms generate embeddings that fail to properly preserve the network structure, or lead to unstable representations due to random processes (e.g., random walks to generate context) and, thus, cannot generate to multi-graph problems. %Therefore, on one side of the spectrum, such algorithms seem to be suitable for tasks involve a single graph, like node classification. However, on the other side, these algorithms cannot fit multi-graph problems, such as graph similarity, and network alignment. 
In this paper, we propose RECS, a novel, stable graph embedding algorithmic framework. RECS learns graph representations using connection subgraphs by employing the  analogy of graphs with electrical circuits. It preserves both local and global connectivity patterns, and addresses the issue of high-degree nodes. Further, it exploits the strength of weak ties and meta-data that have been neglected by baselines. The experiments show that RECS outperforms state-of-the-art algorithms by up to $36.85$\% on multi-label classification problem. Further, in contrast to baselines, RECS, being deterministic, is completely stable.  
\end{abstract}
\section{Introduction}
\label{sec:intro}
%\reminder{this paragraph is too long!} 
%Real-world information networks (a.k.a. graphs), such as social networks, biological networks, co-authorship networks, and language networks are ubiquitous. Further, the large size of networks \textemdash ~millions of nodes and billions of edges~\textemdash~ and the massive amount of information they convey~\cite{easley2010networks}, have led to a serious need for efficient and effective network mining techniques. 

\hide{\reminder{1. maybe add a figure? Saba: can we refer to figure 1 at the end of intro section?}}Conventional graph mining algorithms~\cite{goyal2017graph} have been designed to learn a set of hand-crafted features that best performs to conduct a specific downstream task; i.e., link prediction~\cite{liben2007link}, node classification~\cite{bhagat2011node}, and recommendation~\cite{yu2014personalized}. However, present research has steered the direction towards a more effective way to mine large-scale graphs: feature learning~\cite{bengio2013representation}. That is, a unified set of features that can effectively generalize over distinct graph mining-related tasks is exploited. To this end, recent research efforts have focused on designing either unsupervised or semi-supervised algorithms to learn node representations. Such efforts have been initiated in the domain of natural language processing (NLP)~\cite{mikolov2013distributed,le2014distributed,mikolov2013efficient}, where two word2vec~\cite{mikolov2013efficient} models have been proposed, namely continuous bag of words (CBOW) and Skipgram. Inspired by the recent advancements in the NLP domain, and the analogy in the context, various algorithms have been developed to learn graph representations~\cite{perozzi2014deepwalk,tang2015line,grover2016node2vec}. However, since real-world networks convey more complex relationships comparing to those emerge in corpora, some recent representation learning algorithms algorithms~\cite{perozzi2014deepwalk,perozzi2016walklets,grover2016node2vec} fail to well-preserve  network structure. This in turn impacts the quality of node  representations, which compromises the performance of downstream processes. In addition, state-of-the-art algorithms share a major stability issue that renders them less robust and applicable, especially for multi-graph problems~\cite{heimann2017generalizing,Heimann18arxiv}. In other words, it seems that while baseline representation learning algorithms strive to preserve similarities among nodes to generate and learn node representations, they fail to maintain similarities across runs of any of the algorithms, even with using the same data set\cite{Heimann18arxiv} (e.g., graph similarity \cite{koutra2013deltacon} and network alignment\cite{bayati2009algorithms}).\newline
\indent The quality of the learned representations is heavily influenced by the preserved local and global structure. Therefore, we need to properly and neatly identify node neighborhood notions. For that, and to the best of our knowledge, we are the first to develop a robust graph embedding algorithm that preserves connectivity patterns unique to undirected and (un)weighted graphs. It employs the concept of network flow represented by connection subgraphs~\cite{faloutsos2004fast}. The connection subgraphs avail the analogy with electrical circuits. That is, a node is assumed to serve as a voltage source, and an edge is assumed to be a resistor, where its conductance is considered as the weight of the edge. Forming a connection subgraph allows us to: (1) Concurrently capture the \textit{node local and global connections}, (2) Account for the node degree imbalances by downweighing the importance of paths through high-degree nodes (hops), (3) Take into account both low- and high-weight edges; and (4) Account for \textit{metadata} that is largely being neglected by existing embedding algorithms. 
%The formation of connection subgraphs is divided into two phases: 1) Distance-driven, where we avoid long paths along which information is more likely to be lost, and account for high-degree nodes issue, and 2) Flow-driven, where we account for \textit{metadata} flow among nodes and has been neglected by existing emedding algorithms. 
 To summarize, our contributions are:
\vspace{-0.3cm}
\begin{enumerate}
	\item \textbf {Flow-based Formulation.} We propose a graph embedding approach that robustly-preserves network local and global structure by leveraging the notion of network flow to produce approximate but high-quality connection subgraphs between pairs of non-adjacent nodes in  undirected and (un)weighted large-scale graphs. We use the formed connection subgraphs to identify the node neighborhoods and not restrict ourselves just to one- or two-hop neighbors.
	%\item \textbf {Tailored Objective.} We optimize an objective function that is tailored to obey with network structure and account for node-degree distribution.
	%\item \textbf {Network Structure.} We more carefully capture the local and global network structure comparing to either shallow models adopted the notion of random walks, or deep models employed multilayer neural networks.
	%\item \textbf {Degree Distribution.} We address the issue of high-degree nodes in attaining connection subgraphs that represent proximities among non-adjacent nodes. 
	%\item \textbf {Robust Proximity Measure.} We here propose to robustly capture proximities among ubiquitous non-adjacent nodes in large-scale graphs.
	\item \textbf {Algorithm for Stable Representations.} Contrary to all state-of-the-art methods, which involve randomness, reflected on the embeddings and their quality, 
	our proposed robust graph embedding, titled \method, produces consistent embeddings across independent runs.
	\item \textbf{Experiments.} We extensively  evaluate \method empirically, and we demonstrate that it outperforms the state-of-the-art algorithms in two aspects. (1) Effectiveness: \method outperforms state-of-the-art algorithms by up to $36.85$\% on multi-label classification problem, and (2) Robustness: in contrast to baseline algorithms, experiments show that \method is completely stable by performing a per dimension comparison of embeddings obtained from two runs of the same algorithm using an identical data set.
\end{enumerate}

%\reminder{If we are running low on space, the following paragraph is a bit tautological typically so we can omit it :)}
%The rest of the paper is structured as follows. In Section~\ref{sec:related}, we briefly highlight conventional and recent representation learning methods. We formally define our problem statement in Section~\ref{sec:problem}. In Section~\ref{sec:method}, we present the details of our proposed algorithm; \method. In Section~\ref{sec:Evaluation}, we evaluate the effectiveness and robustness of \method on node classification task using data sets from various domains. We finally conclude our work and highlight future directions in Section~\ref{sec:conclusions}.

\section{Related Work}\label{sec:related}
\noindent \textbf{Representation Learning.} Recent work in network representation learning has been largely motivated by the new progress in natural language processing (NLP) domain~\cite{mikolov2013distributed,le2014distributed,mikolov2013efficient}, due to the existing analogy among the two fields, where a network is represented as a document. One of the NLP leading advancements is rooted to %to the word2vec %two developed
%models~\cite{mikolov2013efficient}, namely, continuous bag of words (CBOW) and
the SkipGram model, due to %. The SkipGram  model has been more widely adopted by virtue of
its efficiency in scaling to large-scale networks. %In a nutshell, it learns continuous distributed feature representations of words in a corpus by maximizing the probability of observing the context words given the representation of a target word. The underlying idea is that the words that usually appear in similar contexts tend to have similar meaning, thus have similar representation vectors, which in turn leads to embedding them close to one another in the low-dimensional continuous vector space.
However, merely adopting the SkipGram model for graph representation learning seems to be insufficient in capturing local and global connectivity patterns~\cite{perozzi2014deepwalk,tang2015line,grover2016node2vec}, because of the sophisticated connectivity patterns that emerge in networks, but not in text corpora. 
%For that, we propose to preserve linear and non-linear proximities while generating neighborhood notions and before even getting learned by the SkipGram model. 
%Recent proposed algorithms, which adopted SkipGram model are incapable to satisfactorily and properly capture nodes neighborhood notions in a network~\cite{perozzi2014deepwalk,tang2015line,grover2016node2vec}. %which is completely different from what corpora used to have.
Specifically, DeepWalk~\cite{perozzi2014deepwalk}, for instance, employs small truncated random walks to approximate the neighborhood of a node in a graph. LINE~\cite{tang2015line} proposes to preserve the network local and global structure using first- and second-order proximities, respectively. A more recent approach, node2vec~\cite{grover2016node2vec}, proposes to preserve graph unique connectivity patterns, homophily and structural equivalence, using biased random walks. Unlike these works, to satisfactorily define the node neighborhood notions, we propose to preserve linear and non-linear proximities while generating neighborhood notions, before being learned by the SkipGram model.

\noindent \textbf{Connection Subgraphs.} There is a significant body of work addressing the problem of finding the relationships between a set of given nodes in a network. For instance, \cite{akoglu13pathways} find simple pathways between a small set of marked nodes by leveraging the Minimum Description Length principle, while~\cite{Tong06centerpiece} defines the center-piece subgraph problem as finding the most central node with strong connections to a small set of input nodes. The work on connection subgraphs~\cite{faloutsos2004fast}, which capture proximity among any two non-adjacent nodes in arbitrary undirected and (un)weighted graphs, is the most relevant to ours. In a nutshell,~\cite{faloutsos2004fast} includes two prime phases: \textit{candidate generation}, and \textit{display generation}. In the \textit{candidate generation} phase, a distance-driven extraction of a much smaller subgraph is performed to generate \textit{candidate} subgraph. At a high level, \textit{candidate} subgraph is formed by gradually and neatly `expanding' the neighborhoods of any two non-adjacent nodes until they `significantly' overlap. Therefore, \textit{candidate} subgraph contains the most prominent paths connecting a pair of non-adjacent nodes in the original undirected and (un)weighted graph. The generated \textit{candidate} subgraph serves as an input to the next phase, i.e., the \textit{display generation}. The \textit{display generation} phase removes any remaining spurious regions in the \textit{candidate} subgraph. The removal process is current-oriented; it aims to add an end-to-end path at a time between the two selected non-adjacent nodes that maximizes the delivered current (network flow) over all paths of its length. Typically, for a large-scale graph, the \textit{display} subgraph is expected to have 20-30 nodes. Connection subgraphs have also been employed for graph visualization~\cite{Rodrigues06}. Our work is the first to leverage connection subgraphs to define appropriate neighborhood notions for representation learning.

\hide{\reminder{what about we add a salesman matrix?} \reminder{From V: Also, since the intro has the essential references, would it make sense to move this section right before the conclusions? - if we have space for salesman table, that would be great too (basically, rows are methods, columns are things that each method can do, and ours has to have checkmarks everywhere :)}}

\section{Proposed Method: \method }\label{sec:method}
In this section, we describe our proposed method, \method, a deterministic algorithm that is capable of preserving local and global---beyond two hops---connectivity patterns. It consists of two main steps: (1) Neighborhood definition via connection subgraphs, and (2) Node representation vector update. We discuss the two steps in~\ref{subsec:step1} and~\ref{subsec:step2}, respectively. We note that \method is deterministic, and thus can be applied to multi-graph problems, unlike previous works~\cite{perozzi2014deepwalk,grover2016node2vec,perozzi2016walklets} that employ random processes, such as random walks.

Our method operates on an (un)weighted and undirected graph $G(\V,\E)$, with $|\V|=n$ nodes and $|\E|=m$ edges. For a given node $u$, we define its 1-hop neighborhood as $\N(u)$ (i.e., set of nodes that are directly connected to $u$).

% which is inspired by the  as shown in Figure~\ref{fig:RECS}. Our algorithm is largely inspired by the connection subgraph algorithm proposed in~\cite{faloutsos2004fast}.

\begin{table}[b!]
%\begin{adjustwidth}{-.5in}{-.5in} 
%\footnotesize
\ssmall
\centering
\captionsetup{justification=centering}
\vspace{-0.5cm}
\caption{Qualitative comparison of the connection subgraph algorithm~\cite{faloutsos2004fast} vs. \method.}
\label{tab:recsVSconnection}
\begin{tabular}{@{}lcc@{}}
\hline
             & \textbf{Connection Subgraph}                 & \textbf{\method}                                          \\ \hline
\textbf{Purpose}      & Node proximity (for only 2 nodes) & Neighborhood definition (for the whole graph) \\
\textbf{Step 1}  & Candidate generation (distance-driven)                & Neighborhood expansion (distance-driven)            \\
\textbf{Step 2} & Display generation (delivered current-driven)                  & Neighborhood refinement (current-driven)                  \\
\textbf{Efficiency}       & Inefficient (for the whole graph)                                & More efficient (for the whole graph)                         \\ 
\textbf{Source}       & $u_i$                                & $\forall \: u \in \V$  \\
\textbf{Target}       & $u_j$                                & Universal sink node $z$                         \\ \hline
\end{tabular}
%\end{adjustwidth}
\end{table}

\subsection{\method - Step 1: Neighborhood Definition}
\label{subsec:step1}

The heart of learning node representations is to obtain representative node neighborhoods, which preserve local and global connections simultaneously. Inspired by \cite{faloutsos2004fast}, we propose to define node neighborhoods by leveraging the analogy between graphs and electrical circuits, and adapting the connection subgraph algorithm (discussed in Section.~\ref{sec:related}) to our setting. In Table~\ref{tab:recsVSconnection}, we give a qualitative comparison of \method and the connection subgraph algorithm~\cite{faloutsos2004fast}, highlighting our major contributions. 

The notion of connection subgraphs is beneficial in our setting, since they allow us to:  (1) Better control the search space; (2) Benefit from the actual flow, meta-data, that is being neglected by state-of-the-art algorithms; (3) Exploit the strength of weak ties; (4) Avoid introducing randomness caused by random/biased walks; (5) Integrate two extreme search strategies, breadth-first search (BFS) and depth-first search (DFS)~\cite{zhou2006breadth}; %, where immediate neighbors and neighbors at increasing distances from the pair of non-adjacent nodes---i.e., source and destination nodes---, are considered, respectively
(6) Address the issue of high-degree nodes; %, that is, a node with high-degree distribution has a low chance to be included in the connection subgraph, as a pair of nodes might be incidentally connected through a high-degree distribution node
and (7) Better handle non-adjacent nodes that are ubiquitous in real-world large-scale graphs. 

\begin{figure*}[t!]
\includegraphics[width=0.9\textwidth]{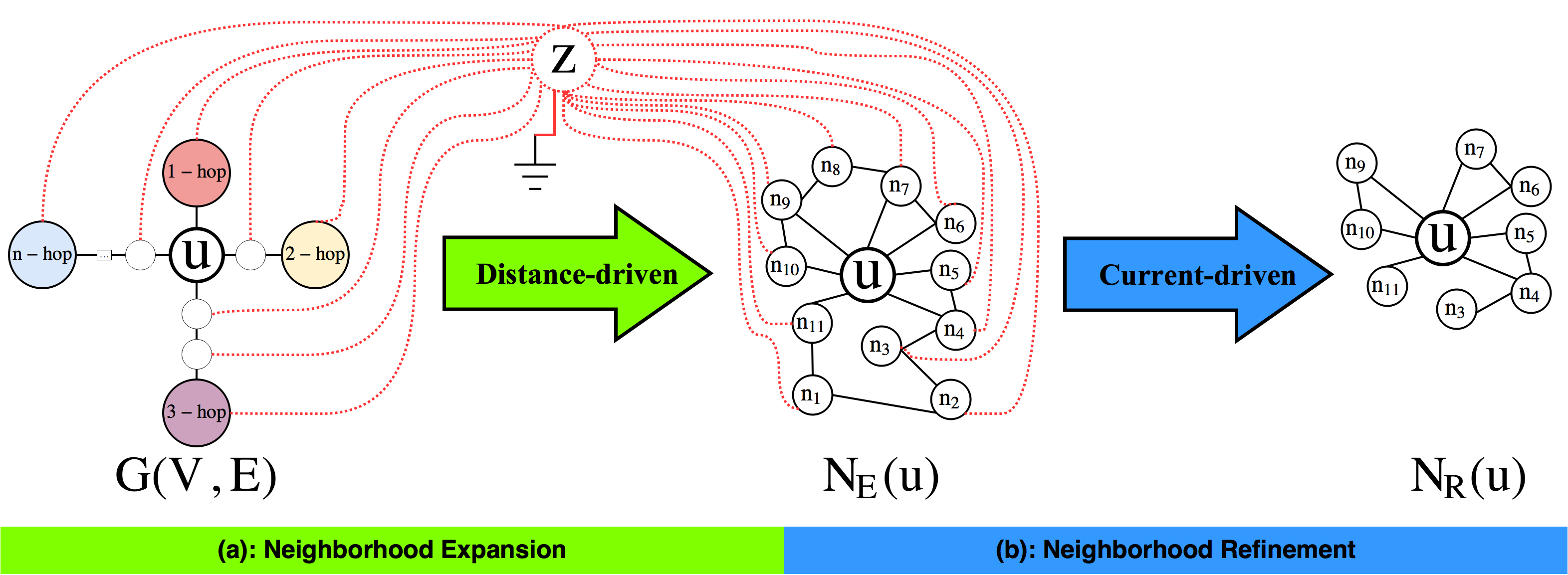}
\centering
\setlength{\abovecaptionskip}{15pt plus 3pt minus 2pt}
%\vspace{-0.5cm}
\caption{A description of \method algorithm neighborhood definition step main phases: (a) Neighborhood expansion of node $u$ through \textit{$n-hop$} neighbors to generate $N_E(u)$ on distance basis. Node $z$ indicates the grounded universal sink node. (b) Neighborhood refinement of $N_E(u)$ to generate $N_R(u)$ on current basis.}
\label{fig:RECS}
\end{figure*}

The neighborhood definition step consists of two phases: (A) \textit{Neighborhood expansion}, and (B) \textit{Neighborhood refinement}. We provide an overview of each phase next, and an illustration in Fig.~\ref{fig:RECS}. The overall computational complexity of \method is $O(\V^2)$.
% According to the analogy with electrical circuits, and by using the connection subgraph algorithm, we define node neighborhoods. 
%The use of a subgraph instead of a single critical path between any two non-adjacent nodes is beneficial for several reasons: (1) Choosing the most important path using a fully automated mechanism is susceptible to errors; (2) Showing a subgraph elevates the probability of having the critical path, if present; and (3) The connection subgraph is better at characterizing the random spread of information in real-world networks than a single path. 

\vspace{0.15cm}
\noindent $\bullet$ \textbf{Phase A: Neighborhood Expansion - $N_E(u)$.}
Given a node $u$, we propose to gradually expand its neighborhood on a distance basis. Specifically, we employ the analogy with electrical circuits in order to capture the distances between $u$ and the other nodes in the network, and then leverage these distances to guide its neighborhood expansion.

% To capture distances  between $u$ and every other node in the network, we first use the analogy with electrical circuits, and then gradually expand $u$'s neighborhood based on proximities. 

\textbf{Graph Construction.} We first construct a modified network $G'$ from $G$  by introducing a universal sink node $z$ (grounded, with voltage $V_z=0)$, and connect all the nodes (except from $u$) to that, as shown in Fig.~\ref{fig:RECS}(a). The newly added edges in $G'$ for every node $v\in \{\V\setminus u\}$ are weighted appropriately by the following weight or conductance (based on the circuit analogy):
\begin{equation} \label{eq:1}
\small
\ C(v,z)=\alpha\sum_{w\in \N(u) \setminus z}C(v,w),
\end{equation}
where $C(v,w)$ is the weight or conductance of the edge connecting nodes $v$ and $w$, $\N(u)$ is the set of 1-hop neighbors of $u$, and $\alpha>0$ is a scalar (set to $1$ for unweighted graphs).

In the modified network $G'$, the distance, or proximity, between the given node $u$ and every other node is defined as:
\begin{equation} \label{eq:2}
  D(u,v)=\begin{cases}
    \log \frac{deg^{2}(u)}{C^{2}(u,v)}, & \text{for } v\in\N(u).\\ 
    \log D(u,c)+ D(c,v), & \text{for } v\notin\N(u),\text{ and } u,v\in\N(c).
    % \log \frac{deg^{2}(u)}{C^{2}(u,c)}+ \log \frac{deg^{2}(c)}{C^{2}(c,v)}, & \text{for } v\notin\N(u),\text{ and } u,v\in\N(c).
  \end{cases}
\end{equation}
where $deg(u)$ is the weighted degree of $u$ (i.e., the sum of the weights of its incident edges), and the distance for non-neighboring nodes $u$ and $v$ is defined as the distance from each one to their nearest common neighbor $c \in \V$. This distance computation addresses the issue of high-degree nodes (which could make `unrelated' nodes seem `close') by significantly penalizing their effects in the numerator.

\begin{figure}[h]
\includegraphics[width=0.5\textwidth]{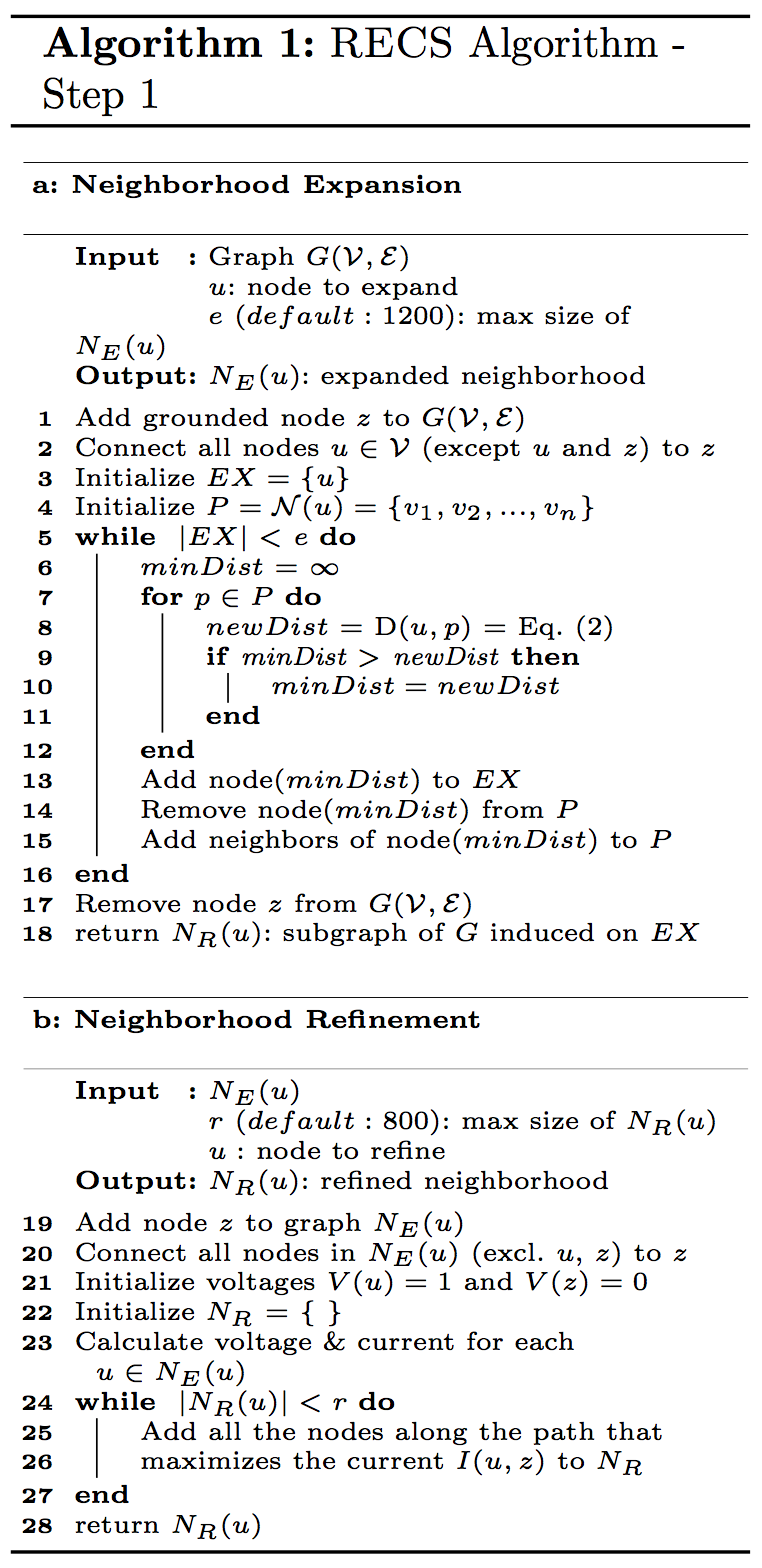}
\centering
\caption*{}
\label{fig:step1}
\end{figure}

\textbf{Distance-based Expansion.} After constructing the circuit-based graph, we can leverage it to expand $u$'s neighborhood. Let $EX$ be the set of expanded nodes that will form the expansion graph $N_E(u)$ (initialized to \{$u$\}), and $P$ be the set of \textit{pending} nodes, initialized to $u$'s neighbors, $\N(u)$. During the expansion process, we choose the closest node to $u$ (except for $z$), as defined by the distance function in Eq.~\eqref{eq:2}. Intuitively, the closer the expanded node $v$ to the source node $u$, the less information flow we lose. Once a node $v$ is added to the expansion subgraph, we add its immediate neighbors to $P$, and we repeat the process until we  have $|EX| = e$ nodes, where $e$ is a constant that represents the desired size of expanded subgraph. We show the \textit{neighborhood expansion} pseudocode in Algorithm 1-a. The procedure of computing the $N_E(u)$ takes $O(\V)$ time. %\reminder{DK: RECS is presented as a 2-step approach... I think we should give both steps as the algorithm...}

\begin{example}
{\it Figure~\ref{fig:step1} shows one example of generating $N_E(u)$ for an undirected, unweighted graph $G$, in which the original edges have conductance (weight) equal to 1, and the size of the expanded neighborhood is set to $e=5$. The conductances for the new edges in $G'$ (red-dotted lines), computed via Eq.\eqref{eq:1}, are shown in Fig.~\ref{fig:step1}-a. Based on the distances between $u$ and every other node, which are defined by Eq.~\eqref{eq:2} and shown in  
Fig.~\ref{fig:step1}-f, the neighborhood of $u$ is expanded on a distance basis. }
%  In order to compute conductance values of edges connecting nodes $1-56 $ with node $z$, we employ Eq.\eqref{eq:1}. We assume $\alpha=1$. The computed conductance values are shown on the red-dotted edges, as shown in Figure~\ref{fig:step1}-a. 
%  After the conductance computations, we compute the distances between node $u$ and the remaining nodes $1-56$ using Eq.~\eqref{eq:2}. Figure~\ref{fig:step1}-f shows the values of the computed distances. As shown in Figure~\ref{fig:step1}(b-e), we expand node $u$'s neighborhood by gradually adding nodes $1, 2, 3$ and $51$, respectively, on a distance basis. Since we assume $e=5$, we stop the expansion after including the four closest nodes to node $u$. 
\end{example}
 
\begin{figure}[h!]
\vspace{-0.8cm}
\includegraphics[width=\textwidth]{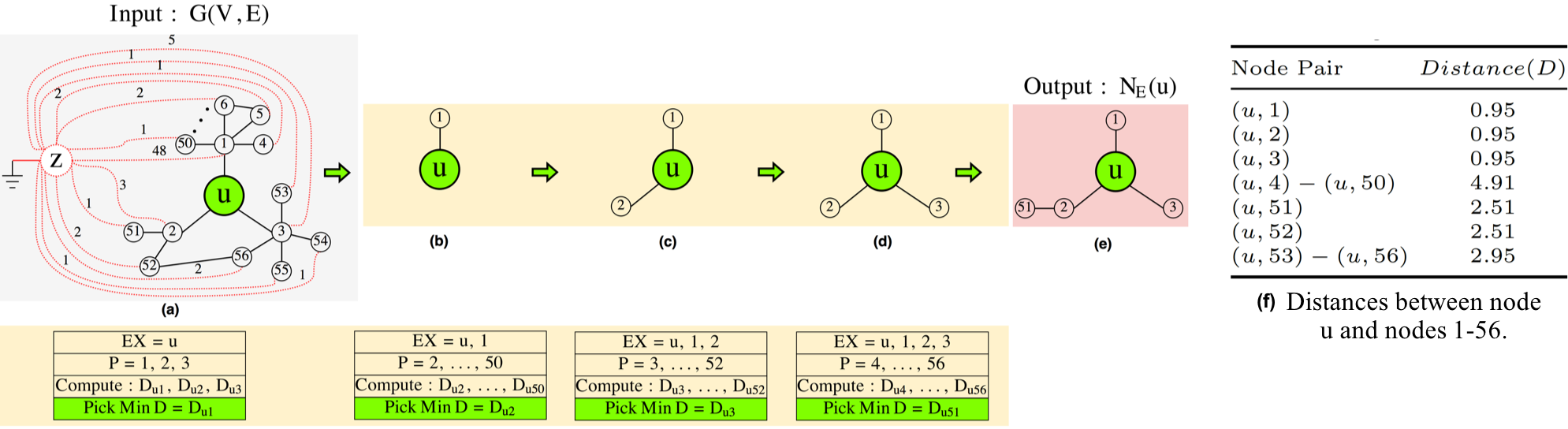}
\centering
\setlength{\abovecaptionskip}{15pt plus 3pt minus 2pt}
\vspace{-0.5cm}
\caption{Neighborhood expansion example.
% (a): Represents the given graph, $G$. The numbers on the red-dotted edges represent the conductance values of edges connecting nodes $1$-$56$ with node $z$. (b-e): Represent the expansion process. To expand node $u$, nodes $1, 2, 3,$ and $51$ are added, respectively, on a distance basis. The closer the neighbor to node, $u$, the higher the probability it has to be added to $N_E(u)$.}
}
\label{fig:step1}
\vspace{-0.7cm}
\end{figure}

\vspace{0.15cm}
\noindent  $\bullet$ \textbf{Phase B: Neighborhood Refinement - $N_R(u)$.} 
% In this phase, as shown in Figure~\ref{fig:RECS}-b, we remove any remaining spurious regions in each \textit{expanded neighborhood} subgraph. 
As shown in Figure~\ref{fig:RECS}-b, the \textit{neighborhood refinement} phase takes an \textit{expanded} subgraph as an input and returns a \textit{refined neighborhood} subgraph as an output, which is free of spurious graph regions. Unlike the previous phase that is based on distances, the \textit{refined} subgraph is generated on a network flow (current) basis.

In a nutshell, in this phase, we first link the nodes of the \textit{expansion} subgraph $N_E(u)$ (except for node $u$) to the previously introduced grounded node $z$. Then, we create the \textit{refined neighborhood} subgraph by adding end-to-end paths from node $z$ to node $u$ one at a time, in decreasing order of total current. The underlying intuition of the \textit{refinement} phase is to maximize the current reaches to node $z$ from the source node $u$. By maximizing the current, we maximize the information flow between the source node $u$ and node $z$, which ultimately serves our goal of including proximate nodes to the source node $u$ in its $N_R(u)$. The process stops when the maximum predetermined \textit{refined} subgraph size, $|N_R(u)|$, is reached. Each time a path is added to the \textit{refined} subgraph, only the nodes that are not already included in the subgraph are added. We use dynamic programming to implement our \textit{refinement} process, which is like a depth first search (DFS) approach with a slight modification.

To that end, we need to calculate the current $I$ flows between any pair of neighbors in the \textit{expanded} subgraph. In our context, $I$ indicates the meta-data or network flow that we aim to avail. We compute the current $I$ flow from source node $s$ to target node $t$ using Ohm's law:
\begin{equation} \label{eq:3}
\small
I(s,t)=C(s,t) \cdot [V(s)-V(t)]
\end{equation}
where the $V(s) > V(t)$ are the voltages of $s$ and $t$, satisfying the \textit{downhill constraint} (otherwise, there would be current flows in the opposite direction). In order to guarantee this satisfaction, we need to sort subgraph's nodes in a descending order, based on their calculated voltage values, before we start current computations. The voltage of a node $s \in \V$ is defined as:
\begin{equation}
\label{eq:4}
  V(s)=\begin{cases}
    \frac{\sum_{v \in \N(s)}V(v)\cdot C(s,v)}{\sum_{v}C(s,v)}, & \forall \text{ nodes } s\neq u,z.\\ 
    1, & s=u.\\
    0, & s=z.
  \end{cases}
\end{equation}
where $C(s,v)$ is the conductance or weight of the edge between nodes $s$ and $v$, as defined in Eq.~\eqref{eq:1}.

\begin{figure}[h!]
\vspace{-0.8cm}
\includegraphics[width=\textwidth]{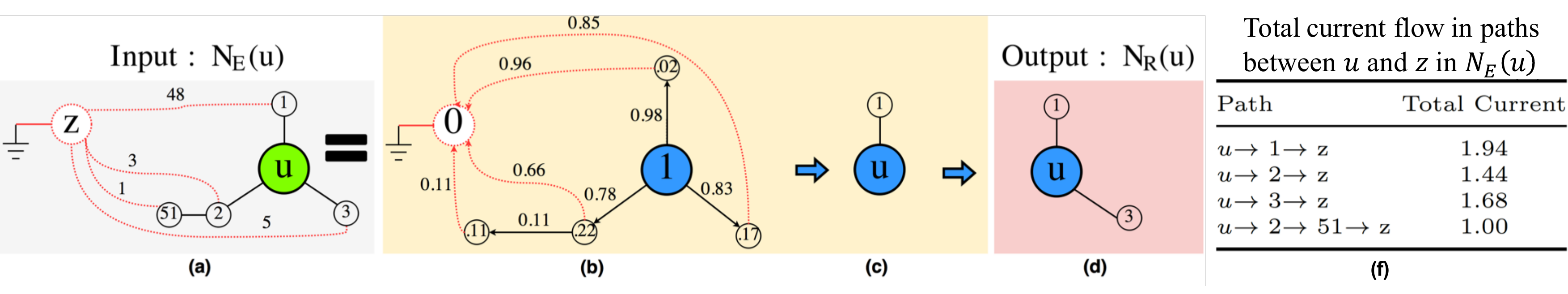}
\centering
\vspace{-0.2cm}
\caption{Neighborhood refinement example. 
%(a): Shows the expanded subgraph, $N_R(u)$, generated by the previous expansion step. The numbers on the red-dotted edges represent the edge conductance values of nodes $1, 2, 3$, and $51$ with node $z$. (b): Shows the computed node voltage values inside the nodes, and the computed currents on the red-dotted and black-solid edges. (c): Shows the addition of the first path connecting node $u$ and node $1$. (d): Shows the addition of the second path connecting node $u$ and node $3$. It also represents the generated $N_R(u)$.}
}
\label{fig:step2}
\vspace{-0.6cm}
\end{figure}

\begin{example}
{\it Given the expanded neighborhood $N_E(u)$ in Example 1, the second phase of \method gradually refines it on a current basis, as shown in Fig.~\ref{fig:step2}. We first compute the voltages by solving the linear system in Eq.\eqref{eq:4}, and include them in the nodes of (b). Then, the current flow of each edge connecting nodes in the expanded neighborhood $N_E(u)$ is computed using Eq.\eqref{eq:3} such that the `downhill constraint' is satisfied (current flowing from high to low voltage), as shown over the red-dotted edges in (b). Given the current values, we enumerate all possible paths between nodes $u$ and $z$, and give their total current flows in (f). The paths are then added in descending order of total current values into $N_R(u)$ until the stopping criterion is satisfied. In (c), we show the first such path. Assuming that the size of the refined neighborhood, $|N_R(u)| = r = 3$ \hide{needs to be less than $r = 3$}, the final neighborhood is given in (d).}
\end{example}

\vspace{0.15cm}
\noindent \textbf{Remark 1: \method neighborhood vs.\ context in baseline methods.} Unlike existing representation learning methods: 
% Our way in identifying node neighborhoods is more concrete and rational comparing to representation learning baseline algorithms for the following reasons: 
(1) We preserve the local and global structure of network by accounting for the immediate neighbors and neighbors at increasing distances of the source node $u$ to identify its neighborhood; (2) We generate neighborhoods on distance and network flow bases; (3) We address the issue of high-node degree distribution; (4) We concurrently identify neighborhoods while maximizing proximity among non-adjacent nodes, which are abundant in real-world networks; and (5) We design our algorithm such that it yields consistent stable representations that suite single and multi-graph problems.      

\vspace{0.1cm}
\noindent \textbf{Remark 2: \method vs.\ connection subgraph algorithm~\cite{faloutsos2004fast}.} It is important to note that the computations of `current' (in \method) and `delivered current' (in~\cite{faloutsos2004fast}) are different. The computation of current is not as informative as delivered current, but is more efficient. The use of delivered current was not a major struggle in~\cite{faloutsos2004fast}, because that algorithm only processes one subgraph. However, we find that it is problematic for generating multiple neighborhoods due to: (1) The large size of the \textit{expanded} subgraph, $|N_E(u)|$; (2) The large size of \textit{refined} subgraph, $|N_R(u)|$ (order of 800), compared to the \textit{display generation} subgraph size capped at $30$ nodes; and (3) The extremely large number of subgraphs (equal to the number of nodes $|\V|=n$) that need to be processed, to ultimately generate node neighborhoods.

\subsection{\method - Step 2: Node Representation Vector Update} 
\label{subsec:step2}

After identifying node neighborhoods in a graph, we aim to learn node representations via the standard SkipGram model~\cite{mikolov2013efficient}. However, since \method yields completely deterministic representations, we avoid the randomness implied by the SkipGram model by using the same random seed every time we employ it. % to learn node representations by optimizing the following objective. 
The Skipgram objective maximizes the log-probability of observing the neighborhood generated during the neighborhood definition step, given each node's feature representation: 

\begin{equation} \label{eq:7}
\small
\underset{f}{\text{max}}\sum_{u\in V}\log(Pr(N_R(u)\mid f(u)),
\end{equation}
where $N_R(u)$ is the refined neighborhood of node $u$, and $f(u)$ is its feature representation.  
Following common practice, we make the maximum likelihood optimization tractable by making two assumptions:\\

\noindent \textbf{Assumption 1 -- Conditional independence.} We assume that the likelihood of observing node $u$'s neighborhood is independent of observing any other neighborhood, given its feature representation $f(u)$:

\begin{equation} \label{eq:8}
\small
Pr(N_R(u)\mid f(u))=\prod_{w\in N_R(u)}Pr(w\mid f(u)) 
\end{equation}
where $w$ represents any node that belongs to node $u$'s refined neighborhood.\\

\noindent \textbf{Assumption 2 -- Symmetry in feature space.} The source node $u$ and any node $w$ in its refined neighborhood $N_R(u)$, have a symmetrical impact on each other in the continuous feature space. Therefore, the conditional probability, $Pr(w \mid u)$, is modeled using the softmax function:

\begin{equation} \label{eq:9}
\small
Pr(w\mid f(u)) =\frac{\exp(f(w)\cdot f(u))}{\sum_{v\in V}\exp(f(v)\cdot f(u))}
\end{equation}

Based on the above two assumptions, we can simplify the objective in Eq.\eqref{eq:7} as follows:

\begin{equation} \label{eq:10}
\small
\underset{f}{\text{max}}\sum_{u\in V}\bigg[-\log\sum_{v\in V}\exp(f(v)\cdot f(u))+\sum_{w\in N_R(u)}f(w)\cdot f(u)\bigg]
\end{equation}

It is important to note that performing such calculations for each node in large-scale graphs is computationally expensive. Therefore, we  approximate the function using negative sampling~\cite{mikolov2013distributed}. We optimize the objective shown in Eq.\ref{eq:10} using stochastic gradient decent.

\hide{\section{Computational Complexity}
We analyze the computational complexity of \method~- \textit{Step 1} as follows:\\

\noindent \textbf{Computing the expanded neighborhood subgraph - $N_E(u)$.} In \method, the procedure of computing the $N_E(u)$ takes $O(\V)$ time.

\noindent \textbf{Computing the refined neighborhood subgraph - $N_R(u)$.} In \method, the procedure of computing the $N_R(u)$ takes $O(|N_E(u)|\times|N_R(u)|)$ time. \reminder{I am not sure if I got this procedure complexity right, because it has a traceback step. The dynamic programming table size is $|N_E(u)|\times|N_E(u)|$.}

\noindent \textbf{\textit{Step 1} overall computational complexity.} In \method, conducting Step 1 takes $O(\V^2)$ time.

}

\section{Experiments}\label{sec:Evaluation}
In this section, we aim to answer the following questions: (\textbf{Q1}) How does \method perform in multi-label classification compared to baseline representation learning approaches? (\textbf{Q2}) How stable are the representations that \method and baseline methods learn? (\textbf{Q3}) How sensitive is \method to its hyperparameters?  Before we answer these questions, we provide an overview of the datasets, and the baseline representation learning algorithms that we use in our evaluation.
%we use to evaluate the efficacy and robustness of the proposed \method algorithm on nontrivial multi-label classification problems.

\vspace{0.1cm}
\noindent \textbf{Datasets.} To showcase the generalization capability of \method over distinct domains, we
use a variety of datasets, which we briefly describe in Table~\ref{my-label1}.

% Table~\ref{my-label1} provides a brief description of evaluation datasets.

\begin{table}[h!]
\small
\centering
\vspace{-0.5cm} 
\caption{A brief description of evaluation datasets.}
\label{my-label1}
{
\setlength{\tabcolsep}{0.7em}
\begin{tabular}{@{}lcccc@{}}
\toprule
\textbf{Dataset}    & \textbf{\# Vertices} & \textbf{\# Edges} & \textbf{\# Labels} & \textbf{\ Network Type} \\ \hline

PPI~\cite{breitkreutz2007biogrid}         & 3,890       & 76,584   & 50        & Biological      \\
Wikipedia~\cite{mahoney2011large}   & 4,777       & 184,812  & 40        & Language      \\
BlogCatalog~\cite{tang2012scalable} & 10,312      & 333,983  & 39        & Social      \\
%DBLP        & 27,199      & 66,832   & 4        & Co-author      \\
CiteSeer~\cite{sen2008collective}    & 3,312       & 4,660    & 6         & Citation      \\
Flickr~\cite{tang2012scalable}    & 80,513       & 5,899,882    & 195         & Social      \\ \hline
\end{tabular}
}
\vspace{-0.3cm}
\end{table}

%\begin{itemize}
%\item \textbf {BlogCatalog~\cite{tang2009relational}:} is a social network of bloggers, where labels indicate blogger interests extracted from meta-data offered by the bloggers.
%\item \textbf {Protein-Protein Interaction (PPI)~\cite{grover2016node2vec}:} is a biological network, where we use a subgraph of a much larger PPI network for Homo Sapiens. The subgraph includes nodes that were able to attain their labels from hallmark gene sets ~\cite{liberzon2011molecular}. The obtained labels indicate biological states.
%\item \textbf {Wikipedia~\cite{grover2016node2vec}:} is a language network for words co-occur with each other. The words appear in the first million bytes of the Wikipedia dump. The labels indicate the Part-of-Speech (POS) tags derived by the Stanford POS-Tagger ~\cite{toutanova2003feature}.
%\item \textbf {DBLP~\cite{perozzi2016walklets}:} is a co-author network for researchers who are working in the field of computer science. The labels represent areas of research researchers interested in. The research areas include data mining (DM), machine learning (ML), database (DB), and information retrieval (IR).
%\item \textbf {CiteSeer~\cite{sen2008collective}:} is a citation network that indicates the connections among publications in the computer science field. The labels represent the research area(s) a paper belongs to. The research areas include ML, DB, IR, artificial intelligence (AI), human-computer interaction (HCI), and Agents.
%\end{itemize}

\vspace{0.1cm}
\noindent \textbf{Baseline Algorithms.} 
We compare \method with three state-of-the-art baselines: {\bf DeepWalk}~\cite{perozzi2014deepwalk}, {\bf node2vec}~\cite{grover2016node2vec}, and {\bf Walklets}~\cite{perozzi2016walklets} \hide{\reminder{why these?}}. The reason why we choose these state-of-the-art methods is the random way they adopt for neighborhood definition using random walks. On the contrary, in \method, we follow a completely deterministic manner, which makes our method applicable for single and multi-graph problems. For all of the methods, we set the number of walks per node to $10$, walk length to $80$, the neighborhood size to $10$, and number of dimensions of the feature representation $d=128$. For node2vec, we set the return parameter $p=1$, and the in-out parameter $q=1$, in order to capture the homophily, and the structural equivalence connectivity patterns, respectively \hide{\reminder{intuition? homophily vs. structural equivalence}}. For Walklets, we set the feature representation scale, $k=2$, which captures the relationships captured at scale $2$.

\vspace{0.1cm}
\noindent \textbf{Experimental Setup.} 
For \method parameter settings, we set the expansion neighborhood subgraph size $|N_E(u)|=1,200$. In order to compare with the baseline methods, we set the refinement neighborhood subgraph size, $|N_R(u)|=800$, and the number of dimensions of the feature representation, $d=128$, in line with the values used for DeepWalk, node2vec, and Walklets. \hide{\reminder{Why do you say that the parameters for \method are in line with the params of the baselines? In what sense are the neighborhoods along these lines?}}

\subsection{Q1. Multi-label Classification}
% In this section, we present our experimental results %of multi-label classification problem and representation learning stability
% we perform using datasets and baseline algorithms described earlier in section~\ref{sec:Evaluation}.\\

% \noindent \textbf{A. Multi-label Classification.} 

\noindent \textbf{Setup.} Multi-label classification is a single-graph canonical task, where each node in a graph is assigned a single or multiple labels from a finite set $\mathcal{L}$. We input the learned node representations to a one-vs-rest logistic regression classifier with L2 regularization. We perform $10$-fold cross validation and report the mean Micro-F1 score results. We omit the results of other evaluation metrics---i.e., Macro-F1 score, because they follow the exact same trend. It is worth mentioning that multi-label classification is a challenging task, especially when the finite set of labels $\mathcal{L}$ is large, or the fraction of labeled vertices is small~\cite{rossi2017deep}.

\noindent \textbf{Results.} In Table~\ref{my-label2}, we demonstrate the performance of \method algorithm and compare it to the three representation learning state-of-the-art methods. Our results are statistically significant with a $p$-value $<0.02$. Overall, \method outperforms or is competitive with the baseline methods, while also having the benefit of generalizing to the multi-network problems that the other methods fail to address. Below we discuss the experimental results by dataset.\\
%\vspace{0.1cm}
\noindent \textbf{PPI:} It is remarkable that using various percentages of labeled nodes, \method outperforms all the baselines. For instance, \method is more effective than DeepWalk by $36.85\%$ when the labeled nodes are sparse ($10\%$), $19.08\%$ for $50\%$ of labeled nodes, and $17.55\%$ when the percentage of labeled nodes is $90\%$. \\
%\vspace{0.1cm}
\noindent \textbf{Wikipedia:} We observe that \method outperforms the three baseline algorithms by up to $10.48\%$ when using $90\%$ of labeled nodes. In the only case where \method does not beat node2vec, it is ranked second.\\
\noindent \textbf{BlogCatalog:} We observe that \method has a comparable or better performance than DeepWalk and Walklets for various percentages of labeled nodes. Specifically, it outperforms DeepWalk by up to $4.55\%$ and Walklets by up to $19.75\%$, when the percentage of labeled nodes is $90\%$. For more labeled nodes, \method achieves similar performance to node2vec.\\
\noindent \textbf{CiteSeer:} Similar to Wikipedia, \method outperforms the state-of-the-art algorithms, and achieves a maximum gain of $7.13\%$ with $90\%$ of labeled nodes. \\
\noindent \textbf{Flickr:} We perceive that \method outperforms the other three baselines by up to $6.79\%$, when using $50\%$ of labeled nodes. \\
\noindent \textbf{Discussion:} From the results, it is evident that \method mostly outperforms the baseline techniques on PPI, Wikipedia, CiteSeer, and Flickr networks, with exceptions, where \method was very close to the best method. This can be rooted in the fact that \method is more capable in preserving the global structure in such networks. On the other hand, although \method has a very comparable performance with node2vec on BlogCatalog dataset, it might be that the \nth{2} order biased random walks of node2vec are slightly more capable in preserving the homophily, and the structural equivalence connectivity patterns in social networks.  

\hide{
\begin{table}[t]
\begin{adjustwidth}{-.5in}{-.5in}  
\ssmall
\centering
\captionsetup{justification=centering}
\caption{Micro-F1 scores for multi-label classification on various datasets. Numbers where \method outperforms other baselines are bolded.}
\label{my-label2}
{%\footnotesize
\begin{tabular}{@{}l|ccc|ccc|ccc|ccc@{}}
\hline
\multirow{2}{*}{Algorithm} & \multicolumn{3}{c|}{BlogCatalog}        & \multicolumn{3}{c|}{PPI}                 & \multicolumn{3}{c|}{Wikipedia}        & \multicolumn{3}{c}{CiteSeer} \\ \cline{2-13} 
                           & 10\% & 50\% & \multicolumn{1}{c|}{90\%} & 10\% & 50\% & \multicolumn{1}{c|}{90\%} & 10\% & 50\% & \multicolumn{1}{c|}{90\%} & 10\% & 50\% & \multicolumn{1}{c}{90\%}  \\ 
\hline
DeepWalk                   &30.12\           &34.28\          &34.83\          &12.35\        &18.23\        &20.39\       &42.33\         &44.57\           &46.19\         &46.56\          &52.01\         &53.32\        \\
node2vec                   & \textbf{34.53}\           & \textbf{36.94}\          & \textbf{37.99}\          &16.19\        &20.64\        &21.75\       &44.38\          & \textbf{48.37} \          &48.85\        & \textbf{50.92}\          &52.49\         &56.72\         \\
Walklets                   &26.90\           &29.09\          &30.41\          &16.07\        &21.44\        &22.10\       &43.69\          &44.68\          &45.17\        &47.89\          &52.73\         &54.83\         \\
\method                       &31.02\           &34.85\          &36.42\          &\bf{16.91}\   &\bf{21.71}\   &\bf{23.97}\  &\bf{45.68}\     &48.10\          &\bf{49.90}\    &48.80\          &\bf{53.36}\         &\bf{57.12}\         \\ \hline
Gain over DeepWalk                   &3.00\           &1.63\          &4.55\          &36.85\        &19.08\        &17.55\       &7.90\         &7.91\           &8.03\         &4.80\          &2.59\         &7.13\        \\
Gain over node2vec           -        &\     -      &\      -    &\    -      &4.41\        &5.16\        &10.19\       &2.92\         &\      -     &2.14\         &\    -      &1.63\         &0.70\        \\
Gain over Walklets                   &15.27\           &19.80\          &19.75\          &5.19\        &1.23\        &8.47\       &4.53\         &7.64\           &10.48\         &1.87\          &1.18\         &4.17\        \\
\hline
\end{tabular}
}
\end{adjustwidth}
\end{table}
}

\begin{table}[t]
% \begin{adjustwidth}{-.1in}{-.1in}  
\centering
\captionsetup{justification=centering}
\caption{Micro-F1 scores for multi-label classification on various datasets. Numbers where \method outperforms other baselines are bolded. By ``G.O.'' we denote ``gain over''.}
\label{my-label2}
\resizebox{\textwidth}{!}{
% {%\footnotesize
\ssmall
\begin{tabular}{@{}l|ccc|ccc|ccc|ccc|ccc@{}}
\hline
\multirow{2}{*}{Algorithm} &   \multicolumn{3}{c|}{PPI}                 & \multicolumn{3}{c|}{Wikipedia} &  \multicolumn{3}{c|}{BlogCatalog}     & \multicolumn{3}{c|}{CiteSeer}   & \multicolumn{3}{c}{Flickr}  \\ \cline{2-16} 
                           & 10\% & 50\% & \multicolumn{1}{c|}{90\%} & 10\% & 50\% & \multicolumn{1}{c|}{90\%} & 10\% & 50\% & \multicolumn{1}{c|}{90\%} & 10\% & 50\% & \multicolumn{1}{c|}{90\%} & 10\% & 50\% & \multicolumn{1}{c}{90\%} \\ 
\hline
DeepWalk    &12.35\        &18.23\        &20.39\       &42.33\          &44.57\            &46.19\      &30.12\             &34.28\            &34.83\            &46.56\              &52.01\         &53.32\     &37.70\   &39.62\  &42.36\ \\
node2vec    &16.19\        &20.64\        &21.75\       &44.38\          & \textbf{48.37} \ &48.85\      & \textbf{34.53}\   & \textbf{36.94}\  & \textbf{37.99}\  & \textbf{50.92}\    &52.49\         &56.72\     &38.90\   &41.39\  &43.91\  \\
Walklets    &16.07\        &21.44\        &22.10\       &43.69\          &44.68\            &45.17\      &26.90\             &29.09\            &30.41\            &47.89\              &52.73\         &54.83\     &38.32\   &40.58\  &42.62\  \\
\method     &\bf{16.91}\   &\bf{21.71}\   &\bf{23.97}\  &\bf{45.68}\     &48.10\            &\bf{49.90}\  &31.02\           &34.85\             &36.42\         &48.80\                 &\bf{53.36}\   &\bf{57.12}\ &\textbf{38.98}\   &\textbf{42.31}\  &\textbf{44.26}\  \\ \hline
G.O. DWalk                  &36.85\     &19.08\     &17.55\          &7.90\        &7.91\        &8.03\       &3.00\         &1.63\           &4.55\             &4.80\           &2.59\          &7.13\    &3.40\   &6.79\  &4.49\ \\
G.O. N2vec                 &4.41\       &5.16\      &10.19\          &2.92\        &-\        &2.14\          &-\         &-\           &-\                      &-\              &1.63\          &0.70\    &0.21\   &2.22\  &0.80\    \\
G.O. Walk                 &5.19\       &1.23\      &8.47\           &4.53\        &7.64\        &10.48\       &15.27\         &19.80\           &19.75\         &1.87\           &1.18\          &4.17\    &1.72\   &4.26\  &3.85\    \\
\hline
\end{tabular}
}
% \end{adjustwidth}
\end{table}

\subsection{Q2. Representation Learning Stability}

\noindent \textbf{Setup.} Surveying the existing node representation learning methods, we perceive that the tasks for which such algorithms are being evaluated on are limited to single graph-related tasks---i.e., prediction, recommendation, node classification, and visualization. Since many tasks involve multiple networks (e.g., graph similarity~\cite{koutra2013deltacon}, graph alignment~\cite{bayati2009algorithms}, temporal graph anomaly detection~\cite{koutra2013deltacon}, brain network analysis for a group of subjects~\cite{fallani2014graph}), we seek to examine the similarity of representations learning approaches to multi-network settings. ~\cite{heimann2017generalizing} states that existing embedding algorithms are inappropriate for multi-graph problems, and attribute this to the fact that different runs of any algorithm yield different representations every time the algorithm is run even if the same dataset is used. To that end, \method is fully deterministic, with the goal of achieving stable and robust outcomes. We evaluate this stability with respect to the following criteria: (1) {\bf Representation Stability}, by verifying the similarity of the learned vectors across different independent runs of the algorithms, and (2) {\bf Performance Stability}, where we use embeddings from different runs in a classification task and we measure the variation in the classification performance. Ideally, a robust embedding should satisfy both criteria.\\

\noindent \textbf{Results.} Here we list the results of the two stability experiments.\\
\noindent \textbf{Representations stability.} Figure~\ref{fig:r_features_p} shows the embeddings of two different runs of each approach against each other for a randomly selected set of nodes. For $d=128$, we visualize the results for three  randomly selected dimensions of node2vec, DeepWalk, and Walklets. For \method, we intentionally choose the same three dimensions randomly selected for each of the baseline methods. In the interest of space, we only show the visualization results of \method using the same three dimensions ($39,55,111$) used for Walklets dataset. The results are equivalent for all the dimensions. If all points fall on (or close to) the diagonal, this indicates stability, which is a desirable attribute of a robust graph embedding. Figures~\ref{fig:r_features_p}(a-c) show that, as expected node2vec, DeepWalk, and Walklets, suffer from significant variation across runs. To the contrary, Figure~\ref{fig:r_features_p}d shows that \method obtain perfectly consistent embeddings across runs, and thus it is robust.

\begin{figure}[t!]
\centering
  \centering
  \includegraphics[width=0.8\linewidth]{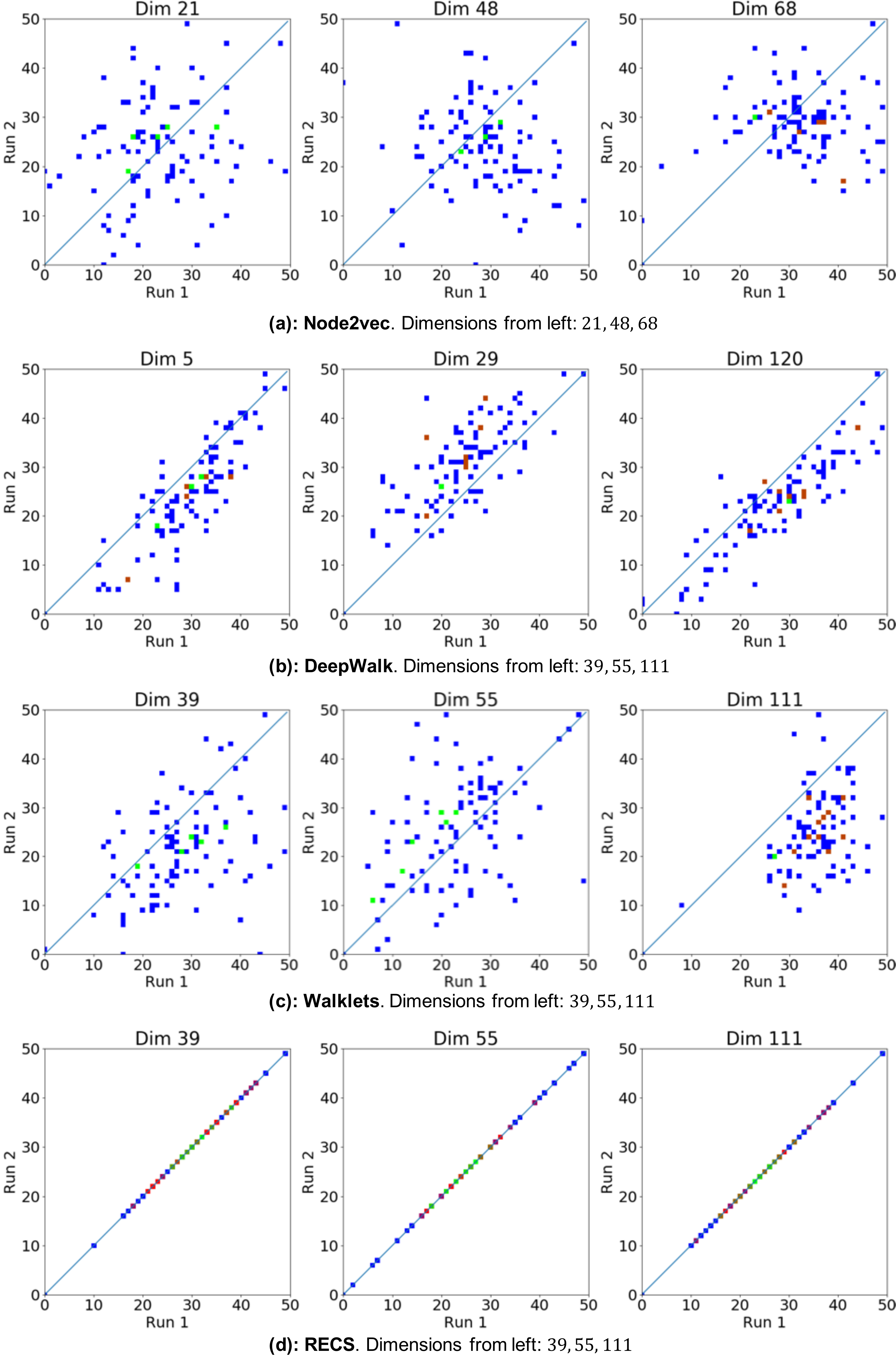}
  \captionof{figure}{PPI data: Comparison of embeddings per dimension for a random sample of $100$ nodes. Node2vec, DeepWalk, Walklets, and \method are run two times. The $x$-axis represents first run representations values, and the $y$-axis represents second run representations values. Three dimensions are selected randomly for each algorithm. The \method-based representations are robust across runs (perfectly fall on a straight line $y=x$), which is not the case for node2vec, DeepWalk, and Walklets. The results are consistent for all the datasets.}
  \label{fig:r_features_p}
\end{figure}%

\hide{\renewcommand*{\thesubfigure}{(\alph{subfigure})}
\begin{figure}[H!]
        \centering
            %\subfigure[DeepWalk. Dimensions from left: 5, 29, 35]{\includegraphics[width=0.8\textwidth]{Figures/PPI_DeepWalk.png}}\label{subfig:1}\quad
            \subfigure[node2vec. Dimensions from left: 21, 97, 53]{\includegraphics[width=0.8\textwidth]{Figures/PPI_Node2vec.png}}\label{subfig:2}\quad
            \subfigure[Walklets. Dimensions from left: 39, 55, 111]{\includegraphics[width=.8\textwidth]{Figures/PPI_Walklets.png}}\label{subfig:3}\quad
            \subfigure[\method. Dimensions from left: 39, 55, 111]{\includegraphics[width=.8\textwidth]{Figures/PPI_SURREAL.png}}\label{subfig:4}
\caption{PPI data: Comparison of embeddings per dimension for a random sample of $100$ nodes. Walklets, node2vec, DeepWalk, and \method are run two times. The $x$-axis represents first run representations values, and the $y$-axis represents second run representations values. Three dimensions are selected randomly for each algorithm. The \method-based representations are robust across runs (perfectly fall on a straight line $y=x$), which is not the case of Walklets, node2vec, and DeepWalk. The results are consistent for all datasets. \reminder{re-run \method using same dimensions of Walklets}}
\label{fig:r_features_p}
\end{figure}
}

\begin{figure}[t!]
\centering
  \centering
  \includegraphics[width=0.7\linewidth]{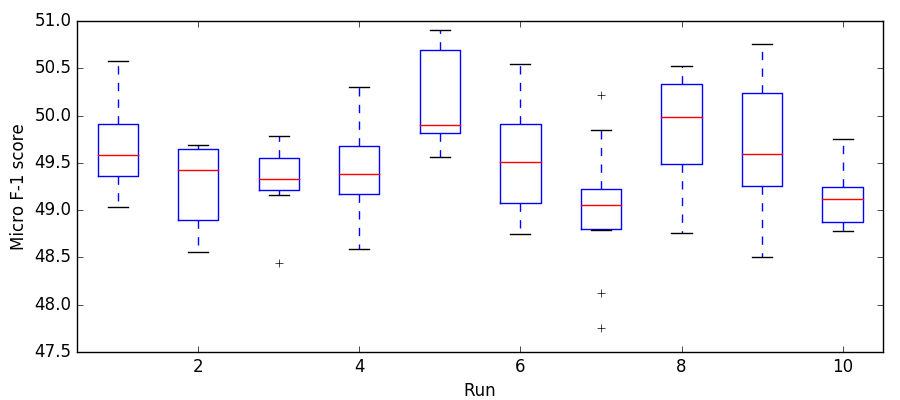}
  \captionof{figure}{Micro-F1 score Boxplots of node2vec on the Wikipedia dataset. Unstable embeddings across multiple runs can statistically impact the performance of the classification task.}
  \label{fig:box}
\end{figure}

\noindent \textbf{Performance stability.} The literature in representation learning has routinely overlooked the effect of instability/randomness of the learned representations and its effect on performance of downstream tasks. In other words, our performance stability hypothesis states that \textit{in addition to representation quality, representation stability also matters}. For that, we run node2vec, the approach that sometimes outperformed ours in the classification task, 10 times using evaluation datasets to see if unstable embeddings can statistically impact multi-classification task performance. For all the datasets, we get a $p$-value $<0.05$. Specifically for Wikipedia, $p$-value $=0.00$, which we show in Figure~\ref{fig:box}. Therefore, in addition to the learned representations quality, performance can be compromised by the learned representations instability. This emphasizes the significance of robustly learning node representations. 

\subsection{Q3. Parameter Sensitivity} 

For sensitivity analysis, we use the Wikipedia dataset with $50\%$ labeled nodes. We perform the following three experiments:\\

\noindent \textbf{Size of the expansion neighborhood subgraph $|N_E(u)|$.} First; we demonstrate the impact of varying the size of the expanded neighborhood, $|N_E(u)|$, in a multi-label classification problem. Therefore, we run \method by varying the size of $N_E(u)$ from $600$ to $1,800$ nodes in $200$ increments. We limit the size of the refined neighborhood, $|N_R(u)|=400$. Figure~\ref{fig:sensitivityd}-a shows the Micro-F1 score results. We observe that by increasing the size of $N_E(u)$, the corresponding Micro-F1 score increases up to a certain limit ($|N_E(u)|=1,000$), while it starts to decrease afterwards. This can be attributed to the fact that enlarging the $N_E(u)$ to more than $1,000$ introduces noise to the generated neighborhood, which ultimately compromises the performance.\\

{\begin{figure}[h!]
\centering
  \centering
  \includegraphics[width=\linewidth]{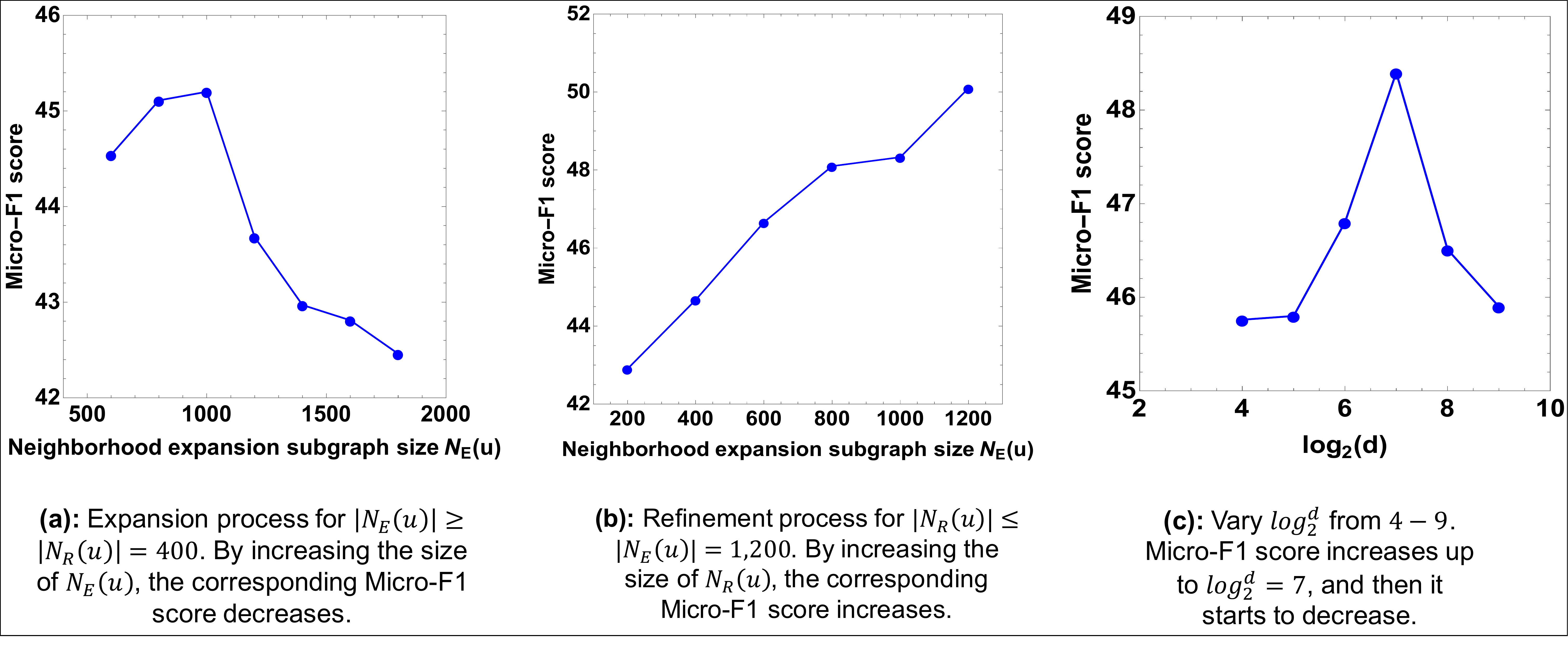}
  \captionof{figure}{Performance sensitivity analysis of \method.}
  \label{fig:sensitivityd}
\end{figure}%

\noindent \textbf{Size of the refinement neighborhood subgraph $|N_R(u)|$}. Fixing the size of expanded neighborhood, $|N_E(u)|=1,200$, we now examine the impact of altering the size of the refined neighborhood, $|N_R(u)|$, in a multi-label classification problem. For that, we run \method, while varying the size of $|N_R(u)|$ from $200$ to $1,200$ nodes in $200$ increments. We set the size of the expanded neighborhood, $|N_E(u)|=1,200$. Figure~\ref{fig:sensitivityd}-b shows the Micro-F1 results. We observe that increasing the $|N_R(u)|$ is accompanied by an increase in the Micro-F1 score. This is rooted in the fact that enlarging the $|N_R(u)|$ includes more useful information in the refined neighborhoods, which SkipGram model~\cite{mikolov2013efficient} leverages to learn and update the node representations. \\

\noindent \textbf{Number of dimensions $d$}. Fixing the sizes of the expanded subgraph, $|N_E(u)|=1,200$, and the refined subgraph, $|N_R(u)|=800$, we demonstrate the impact of varying the representation number of dimensions, $d$, in a multi-label classification problem. For that, we run \method, while varying $\log_2 d$ from $4$ to $9$. Figure~\ref{fig:sensitivityd}-c shows the Micro-F1 results. We note that the Micro-F1 score constantly increases by increasing $\log_2 d$ up to $7$, which corresponds to $d=128$, while it starts to drop afterwards. We root this in the fact that using higher number of dimensions could introduce unrelated dimensions to the representation space, which eventually impact the performance.

\hide{\renewcommand*{\thesubfigure}{(\alph{subfigure})}
\begin{figure} [t!]
\centering
        \centering
           \subfigure[Expansion process for $|N_E(u)|\geq|N_R(u)|=800$. By increasing the size of $N_E(u)$, the corresponding Micro-F1 score decreases.]{
                \includegraphics[width=0.30\textwidth]{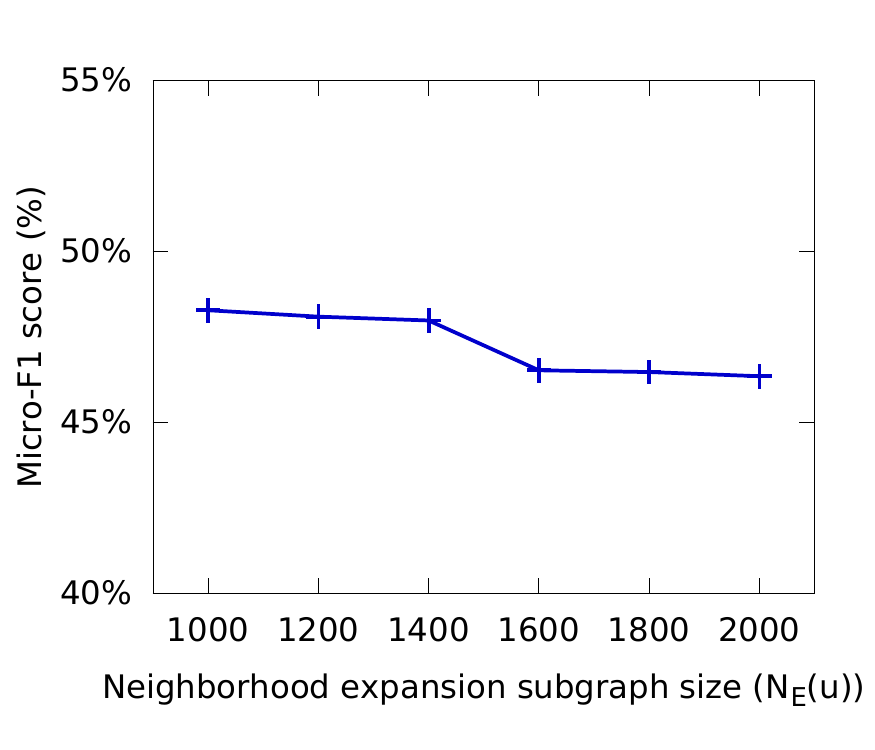}
        }\label{subfig:sensitivity1}\quad \quad
        \subfigure[Refinement process $|N_R(u)|\leq|N_E(u)|=1200$. By increasing the $N_R(u)$ size, the corresponding Micro-F1 score increases.]{
                \includegraphics[width=0.30\textwidth]{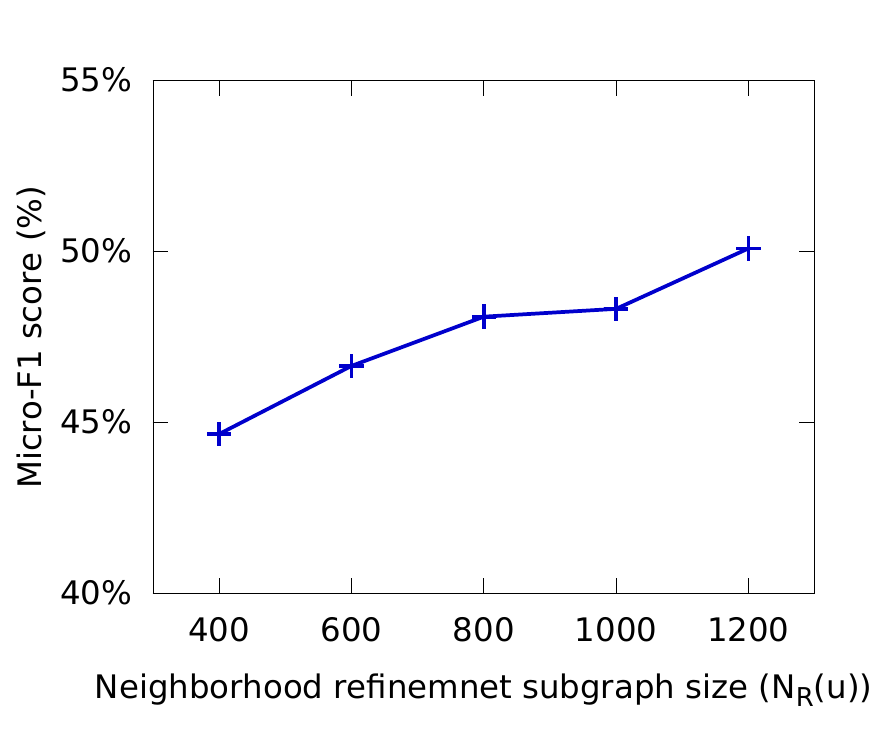}
        }\label{subfig:sensitivity2}
        \subfigure[Vary $\log_2 d$ from $4$ - $9$. Micro-F1 score increases up to $\log_2 d=7$, and then it starts to decrease.]{
                \includegraphics[width=0.30\textwidth]{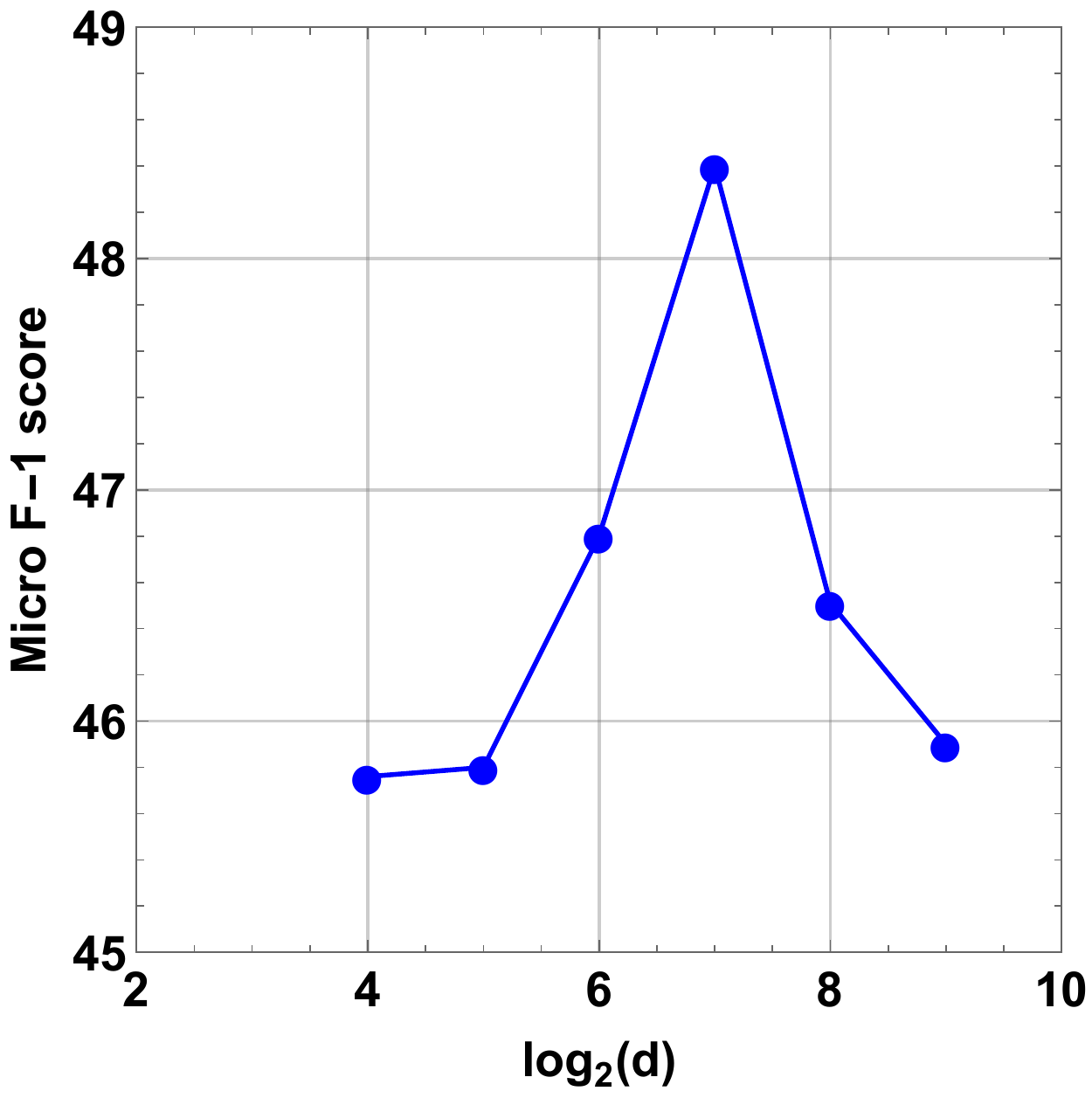}
        }\label{subfig:sensitivity3}
        \captionof{figure}{Performance sensitivity analysis of \method.}
        \label{fig:sensitivity}\quad \quad
\end{figure}
}

\section{Conclusion}
\label{sec:conclusions}
We propose a novel and stable representation learning algorithm; \method, using connection subgraphs. \hide{\method includes two steps: 1) Neighborhood definition using connection subgraphs, which represents the heart of our contribution, and 2) Representation vector update.} %using the established SkipGram model.
%In essence, our contribution lies in the first step, where we generate nodes neighborhood using connection subgraphs. %For that, we divide the neighborhood generation process into two phases: 1) Neighborhood expansion, where we extract a much smaller subgraph from original graph to capture proximity between any two non-adjacent nodes in the graph on distance basis, and 2) Neighborhood refinement, where we do extra pruning to the expanded subgraph in previous phase using information flow (current) basis. It is worth mentioning that despite the fact that connection subgraphs are designated to attain proximity among two non-adjacent nodes, they are still capable to address local connectivity between the pair of such nodes. This can be rooted to the nature of neighborhood generation process.
In contrast to representation learning baseline algorithms, \method generates entirely deterministic representations, which makes it more appealing for single- and multi-graph problems.   %such a formation process renders \method the ability to generate stable consistent representations each time we run the algorithm.
We empirically demonstrate \method's efficacy and stability over state-of-the-art approaches. Experiments show that \method is more or as effective as baselines, and is completely stable. In our future work, we will address the interpretability aspect that is not well-examined in the representation learning literature. We will also address the issue of embedding update, especially for a recently-joined node that has no evident connections. This problem is very related to the ``cold-start'' problem in the recommendation systems, where a new user joins the system and we seek external information for him, in order to properly compute his profile. Similarly, we will explore different forms of external context and meta-data for the recently-joined nodes, which can help us address connection sparsity.

%\section{Acknowledgements}
%{\scriptsize
%Research was supported by the National Science Foundation Grant No. XXXXXX. Any opinions, findings, and conclusions or recommendations expressed in this material are those of the author(s) and do not necessarily reflect the views of the funding parties.
%}

%\balance
%\bibliographystyle{plain}
%\bibliographystyle{ACM-Reference-Format}

\bibliographystyle{plain} 
\bibliography{./BIB/vagelis_refs}  
\end{document}